\documentclass[conference]{IEEEtran}
\IEEEoverridecommandlockouts

\usepackage{fancyhdr}

\usepackage{cite}
\usepackage{amsmath,amssymb,amsfonts}
\usepackage{algorithmic}
\usepackage{graphicx}
\usepackage{textcomp}
\usepackage{xcolor}
\usepackage{hyperref}

\def\BibTeX{{\rm B\kern-.05em{\sc i\kern-.025em b}\kern-.08em
    T\kern-.1667em\lower.7ex\hbox{E}\kern-.125emX}}

\begin{document}


\title{Parameter-Efficient Fine-Tuning of Vision Foundation Model for Forest Floor Segmentation from UAV Imagery\thanks{* Corresponding author}}

\author{\IEEEauthorblockN{
		Mohammad Wasil\textsuperscript{1,2,*}, 
		Ahmad Drak\textsuperscript{1,3},
		Brennan Penfold\textsuperscript{1,3},\\
		Ludovico Scarton\textsuperscript{1,2,3},
		Maximilian Johenneken\textsuperscript{1,3},
		Alexander Asteroth\textsuperscript{1,3},
		Sebastian Houben\textsuperscript{1,2,3}
}
\IEEEauthorblockA{
	\textsuperscript{1} Department of Computer Science\\
	\textsuperscript{2} Institute for Artificial Intelligence and Autonomous Systems (A2S)\\
	\textsuperscript{3} Institute of Technology, Resource and Energy-efficient Engineering (TREE)\\
	Bonn-Rhein-Sieg University of Applied Sciences, Sankt Augustin, Germany\\
	\{firstname.lastname\}@h-brs.de
}
}

\maketitle

\fancypagestyle{withfooter}{
	\renewcommand{\headrulewidth}{0pt}
	\fancyfoot[C]{\footnotesize Accepted to the Novel Approaches for Precision Agriculture and Forestry with Autonomous Robots IEEE ICRA Workshop - 2025}
}
\thispagestyle{withfooter}
\pagestyle{withfooter}

\begin{abstract}
Unmanned Aerial Vehicles (UAVs) are increasingly used for reforestation and forest monitoring, including seed dispersal in hard-to-reach terrains. 
However, a detailed understanding of the forest floor remains a challenge due to high natural variability, quickly changing environmental parameters, and ambiguous annotations due to unclear definitions. To address this issue, we adapt the Segment Anything Model (SAM), a vision foundation model with strong generalization capabilities, to segment forest floor objects such as tree stumps, vegetation, and woody debris. 
To this end, we employ parameter-efficient fine-tuning (PEFT) to fine-tune a small subset of additional model parameters while keeping the original weights fixed. 
We adjust SAM’s mask decoder to generate masks corresponding to our dataset categories, allowing for automatic segmentation without manual prompting.
Our results show that the adapter-based PEFT method achieves the highest mean intersection over union (mIoU), while Low-rank Adaptation (LoRA), with fewer parameters, offers a lightweight alternative for resource-constrained UAV platforms. 

\end{abstract}


\section{Introduction}

Unmanned Aerial Vehicles (UAVs) are increasingly utilized for reforestation and forest monitoring efforts \cite{stamatopoulos2024uav-assisted}. For instance, UAV-assisted seeding can mitigate biological challenges by optimizing water availability through precise depth placement and controlled seed spacing \cite{masarei2019factoring}.
Additionally, UAVs can efficiently disperse seeds in hard-to-reach terrain, which can potentially support large-scale reforestation efforts.
However, identifying suitable seed locations presents an additional challenge, as factors such as water content, plant competition, and physical obstacles like woody debris, logs, and tree stumps must be considered. In this work, we focus on obstacle detection using object segmentation methods.

Pre-trained large vision models (LVMs) or vision foundation models become more available for different tasks such as object recognition \cite{dosovitskiy2021an},  semantic segmentation (e.g., via the Segment Anything Model (SAM))\cite{kirillov2023segment}, and Language-Image Understanding \cite{radford2021learning}.  These models were trained on large amounts of images, videos or texts datasets such as ImageNet, CIFAR-100, or WebImageText \cite{radford2021learning}. Nevertheless, the generalization of these models to remote sensing data from UAVs is more challenging due to the complexity of backgrounds, the diversity of scenes \cite{chen2024rsprompter}, and fewer and smaller datasets.  

Fine-tuning the whole parameters of a foundation model is computationally expensive. Moreover, in \cite{hu2022lora}, Hu et al. claim that fine-tuning a portion of parameters results in comparable performance to that of fine-tuning the whole model. The knowledge transfer to downstream tasks is typically done through parameter-efficient fine-tuning (PEFT) methods that aim at introducing small numbers of trainable parameters, known as PEFT modules, to the original and frozen foundation model parameters \cite{houlsby2019parameter}, \cite{chen2022adaptformer}, \cite{wu2023medicalsamadapter}, \cite{hu2022lora}. In this work, we focus on applying parameter-efficient fine-tuning (PEFT) methods for segmenting forest floor data. We use the Segment Anything model (SAM) \cite{kirillov2023segment}, a semantic segmentation model, as the foundation model for segmenting forest floor objects.

Despite strong zero-shot generalization, SAM segmentation performance can degrade on specific dataset \cite{mazurowski2023segment}. 
For example, in \cite{mazurowski2023segment},  Mazurowski et al. claim that SAM performs better on bigger medical objects. 
Furthermore, SAM only provides masks without category information. To address these limitations, we apply parameter-efficient fine-tuning (PEFT) techniques to adapt SAM to our dataset and modify SAM mask decoder such that the number of segmented masks corresponds to the number of categories in our dataset.

\section{Related Work}
In this section, we review parameter-efficient fine-tuning methods, such as adapters and Low-Rank Adaptation (LoRA), which can be employed to adapt the Segment Anything Model (SAM).

\subsection{Segment Anything Model}
The Segment Anything model (SAM) is a large, pre-trained image segmentation model trained on more than 11 million images and 1 billion+ masks and used as the foundation model for several downstream tasks such as medical image analysis \cite{wu2023medicalsamadapter}, crack segmentation in construction \cite{ge2024fine-tuning}, and remote sensing \cite{chen2024rsprompter}. 

SAM consists of three components, namely an image encoder, a prompt encoder and a mask decoder. The image encoder is based on the Vision Transformer (ViT) model \cite{dosovitskiy2021an}. SAM necessitates prompts to guide the model to output masks given an image input. It supports two types of prompts: 1) sparse prompts with points, box and text; 2) dense prompts with masks.


\subsection{Parameter-Efficient Fine-Tuning Methods}

Parameter-efficient fine-tuning (PEFT) techniques allow us to adapt pre-trained foundation models to specific tasks efficiently by adding and training only a small number of additional parameters. The parameters of the foundation model are kept frozen during this process, which leads to a significant reduction in both memory usage and computational expense. Examples of PEFT methods include Adapter \cite{houlsby2019parameter}, \cite{chen2022adaptformer}, \cite{wu2023medicalsamadapter}  and Low-rank Adaptation \cite{hu2022lora}. 

\subsubsection{Adapter}
\begin{figure}[!htbp]
	\centering
	\includegraphics[width=0.65\linewidth]{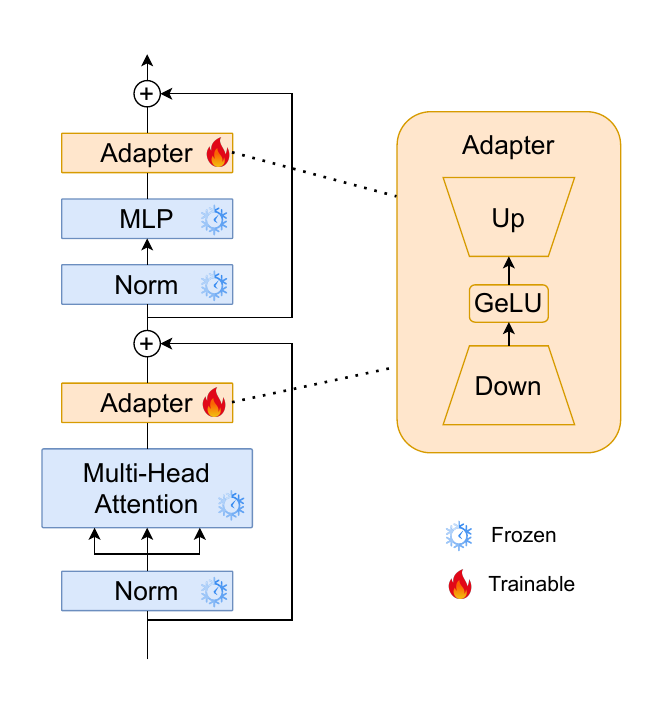} 
	\caption{Adapter injects two serial adapters into each transformer block \cite{houlsby2019parameter}. One adapter is injected after the Multi-Head Attention module, and the other is injected after the MLP module.}
	\label{fig:adapter}
\end{figure}

An adapter tuning method injects new neural network modules into the original foundation model. In \cite{houlsby2019parameter}, Houlsby et al. proposed an adapter method for transfer learning in NLP where two adapter layers are injected into each Transformer layer: one adapter is injected after the multi-head attention and two feed-forward layers as shown in Fig. \ref{fig:adapter}. The adapter module is a bottleneck neural network architecture where the $d$-dimensional features are mapped into a smaller dimension and then projected back to $d$-dimensional output features.

\begin{figure}[!htbp]
	\centering
	\includegraphics[width=0.7\linewidth]{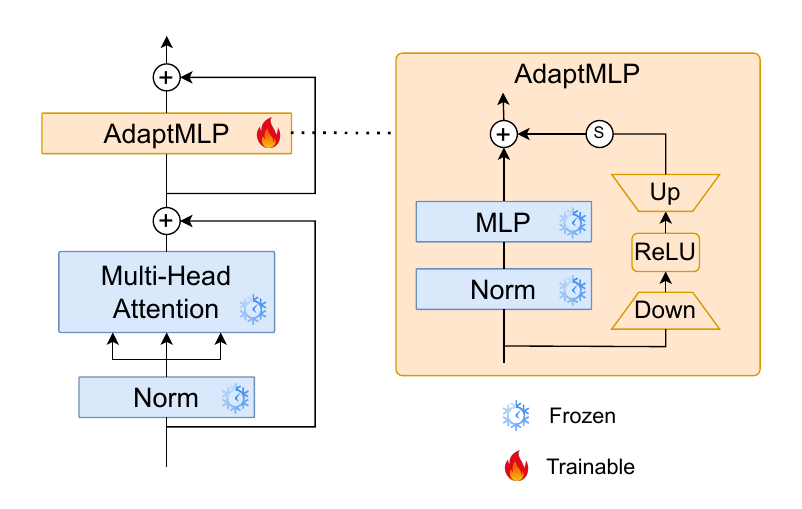} 
	\caption{AdaptFormer introduces a bottleneck adapter alongside with a MLP layer named AdaptMLP \cite{chen2022adaptformer}.}
	\label{fig:adaptformer}
\end{figure}

In \cite{chen2022adaptformer}, Chen et al. introduced a more efficient version of adapter layers by introducing a single residual adapter layer per transformer block. The residual adapter module is inserted alongside the MLP module as depicted in Fig. \ref {fig:adaptformer}. The module is called AdaptMLP that also introduces a scale factor $s$ to the residual connection. The authors claim that the parallel design of adapters in AdaptMLP gives better performance than serial adapters.
In \cite{wu2023medicalsamadapter}, Wu et al. combined the original adapter with AdaptMLP. The first adapter is added after the multi-head attention layer and before the layer norm and the second adapter is added in parallel to the MLP layer.

\subsubsection{Low-Rank Adaptation}
In \cite{hu2022lora}, Hu et al. proposed the Low-Rank Adaptation (LoRA) method, which fine-tunes low-rank matrices for downstream tasks. Their approach is based on the hypothesis that full-rank matrices are not required for downstream tasks, making it unnecessary to fine-tune the entire foundation model.

\begin{figure}[!htbp]
	\centering
	\includegraphics[width=0.8\linewidth]{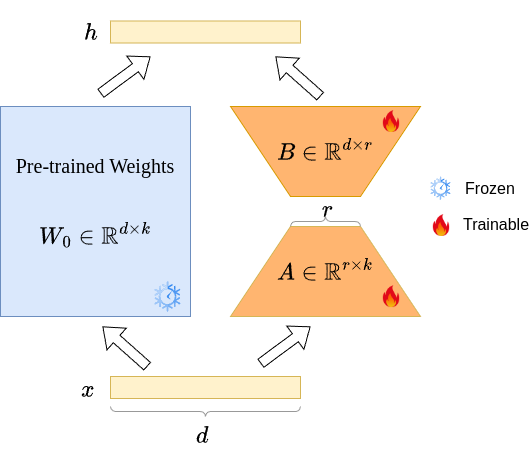} 
	\caption{LoRA architecture with a trainable matrix product BA \cite{hu2022lora}, where the original weight matrix remains frozen. The final embedding $h$ is obtained by element-wise summation of the outputs from the pre-trained weights and the LoRA module.}
	\label{fig:lora}
\end{figure}

LoRA introduces two additional trainable matrices in parallel to the foundation model as shown in Fig. \ref{fig:lora}. Given a pre-trained weight $W_0 \in \mathbb{R}^{d \times k}$, a bottleneck LoRA module $\Delta W$ is introduced, such that the output embedding $h$ becomes: 

$$ h = W_0 x + \frac{\alpha}{r} \Delta W x = W_0 + \frac{\alpha}{r}  B A x $$

where $h$ represents the element-wise sum of the output from the LoRA matrices and the original weight, $B \in \mathbb{R}^{d \times r} $, $A \in \mathbb{R}^{r \times k}$, $\alpha$ is the scaling factor, and $r \ll \min (d,k)$ is the rank. $W_0$ is frozen and only $\Delta W$ is trainable. The weight of $A$ is initialized with Gaussian and $B$ is initialized to zero.

LoRA can be applied to weight matrices in the Transformer self-attention module $W_q, W_k, W_v, W_o$. However, it was found that adapting $W_q$ and $W_v$ gives the best performance, while adapting the weight matrix $W_q$ or $W_k$ results in significantly lower performance \cite{hu2022lora}. The MLP modules, however, are frozen when training the downstream tasks.

\section{Methodology}

This section outlines the methodology for adapting SAM image encoder using adapters and LoRA, as well as modifying SAM's mask decoder to generate output masks corresponding to the number of categories in the dataset.

\subsection{Image Encoder}
\label{subsec:image_encoder}
\begin{figure}[!htbp]
	\centering
	\includegraphics[width=\linewidth]{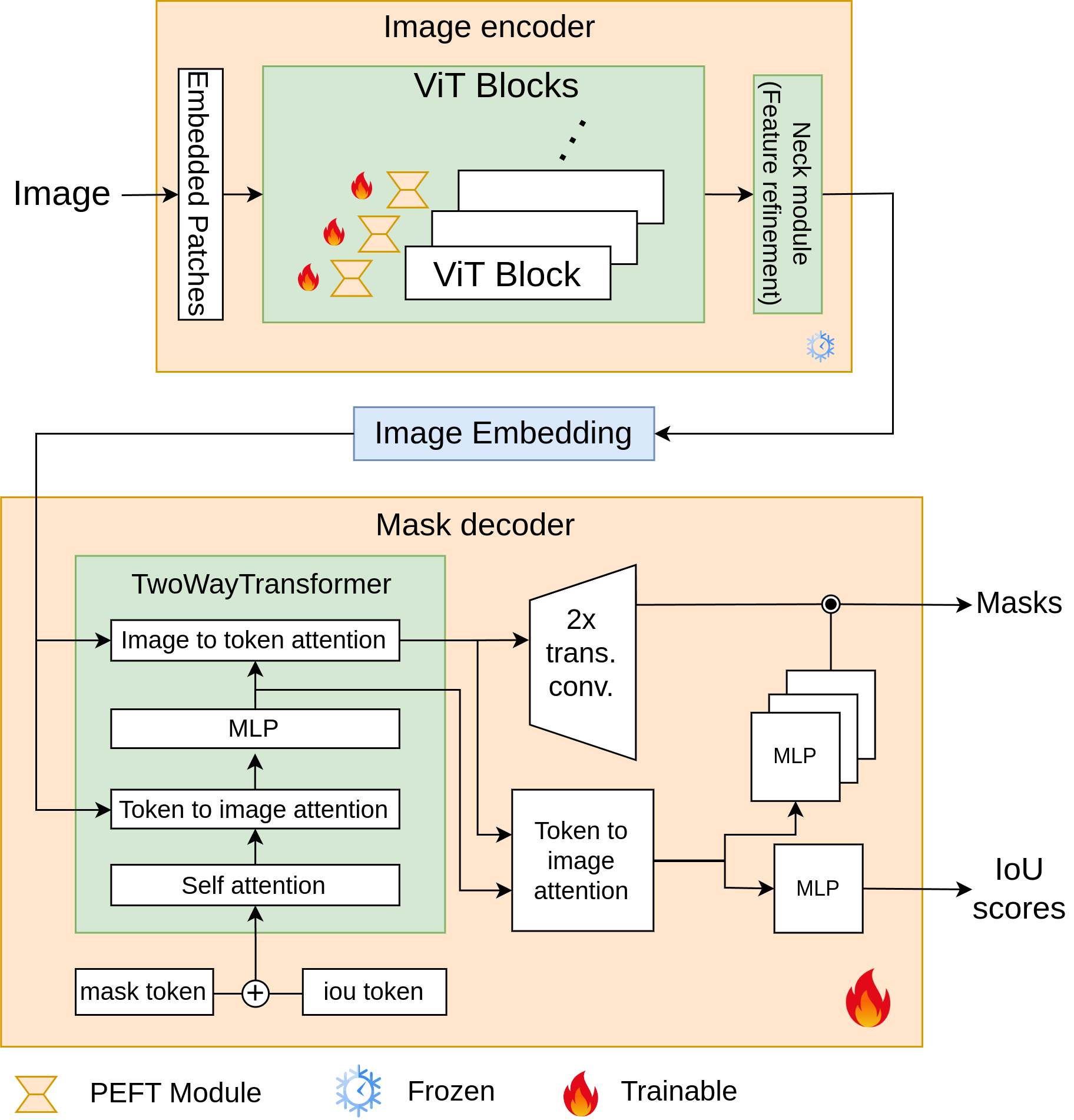} 
	\caption{Modified Segment Anything architecture with PEFT modules. The image encoder is frozen and the PEFT modules and the whole mask decoder parameters are trainable.}
	\label{fig:sam_peft}
\end{figure}

SAM consists of an image encoder, prompt encoder and mask decoder. There are three variants of SAM models, namely SAM ViT-B, SAM ViT-L, and SAM ViT-H. Each of these variants is based on a different ViT model, resulting in different numbers of transformer modules. Specifically, SAM ViT-B, ViT-L, and SAM ViT-H employ 12, 24, and 32 transformer blocks respectively.  These variations result in different numbers of PEFT modules. 
In this work, we select the ViT-H backbone due to its superior performance \cite{kirillov2023segment}.
As shown in Fig. \ref{fig:sam_peft}, the original image encoder parameters remain frozen, while the PEFT modules integrated into each ViT block are trainable. The resulting image embedding is then passed to the mask decoder.

\subsection{Mask Decoder}
The mask decoder in SAM takes the image embedding and prompt tokens, which are generated by the prompt encoder to produce output masks. The prompt encoder translates user inputs such as points, boxes, text, and masks into a unified, high-dimensional prompt embedding. Specifically, point, box and text prompts generate a prompt embedding, whereas mask prompts generate dense embeddings. Then, the mask decoder takes the prompt tokens, the output tokens, and the combined dense and image embedding as inputs and maps them to a mask. In cases where no explicit mask prompts are provided, the mask decoder relies on a learned no-mask embedding.


In this work, we focus on scenarios where no prompts are required because the objects to be segmented are predetermined. Therefore, the prompt and the dense embedding are not set, and we rely solely on the image embedding and the output tokens as inputs to the mask decoder as illustrated in Fig. \ref{fig:sam_peft}. This simplification reduces computational overhead during training and inference. The mask decoder employs a cross-attention mechanism between the image embedding and output tokens. Firstly, a two-way transformer module updates both the image embedding and the tokens. Secondly, the updated image embedding is then up-sampled by a factor of four using convolutional layers, after which the tokens attend to the up-sampled embedding once more. These tokens are subsequently passed through a three-layer MLP, and the outputs are combined with the up-scaled image embedding via a dot product. Finally, this point-wise product generates high-resolution masks with the size of $B \times N \times H \times W$, where  \( B \), \( N \), \( H \), and \( W \) represent the batch size, number of classes, height and width respectively. Unlike the image encoder, where only PEFT modules are trained, we fine-tune the entire mask decoder parameters

\section{Experimental Results}

In this section, we discuss the dataset, experimental setup, results and discussion.

\subsection{Dataset}

\begin{figure}[!htbp]
	\centering
	\includegraphics[width=\linewidth]{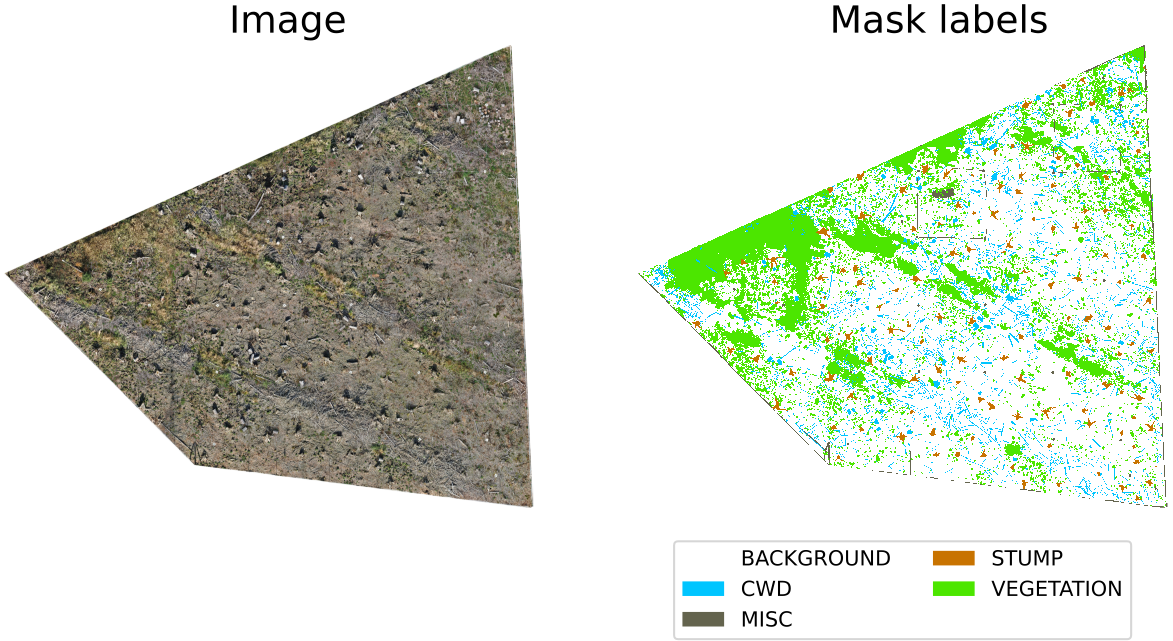} 
	\caption{Example of the orthomosaic image from field-D (left), and its corresponding mask labels (right), featuring coarse woody debris (CWD), miscellaneous (MISC), tree stumps (STUMP), and vegetation.}
	\label{fig:Garrulus_dataset_and_mask_labels}
\end{figure}

This study focuses on the Garrulus dataset, a comprehensive remote sensing and forest floor dataset collected using an Unmanned Aerial Vehicle (UAV) over the Arnsberg Forest in Germany. The surveyed area is divided into several sections, namely field-A, field-B, field-C, and field-D. However, this work specifically focuses on field-D that covers a 0.3 hectare post-harvest area. The field was then reconstructed into a geo-referenced RGB orthomosaic with a high spatial resolution of approximately 10 cm per pixel as shown in Fig. \ref{fig:Garrulus_dataset_and_mask_labels}. Annotations were performed using the Computer Vision Annotation Tool (CVAT) \footnote{https://www.cvat.ai}, with four object classes labeled: coarse woody debris (CWD), tree stumps (STUMP), vegetation, and MISCELLANEOUS or MISC for the ground sampling point marker class, as illustrated in Fig. \ref{fig:Garrulus_dataset_and_mask_labels}. We plan to publish the Garrulus dataset in a separate publication once all relevant fields such as field-A, field-B and field-C have been fully annotated.

\begin{figure}[!htbp]
	\centering
	\includegraphics[width=0.7\linewidth]{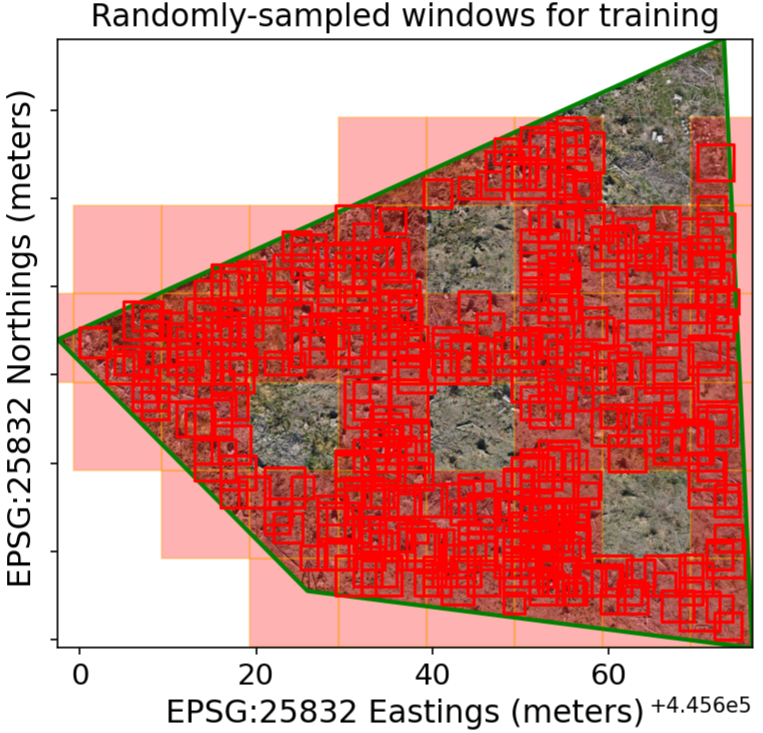} 
	\caption{Example of randomly sampled windows within the training boundary (green border). Tiles without sampled windows are reserved for testing.}
	\label{fig:randomly_sampled_windows_for_train_set}
\end{figure}

\begin{figure}[!htbp]
	\centering
	\includegraphics[width=0.7\linewidth]{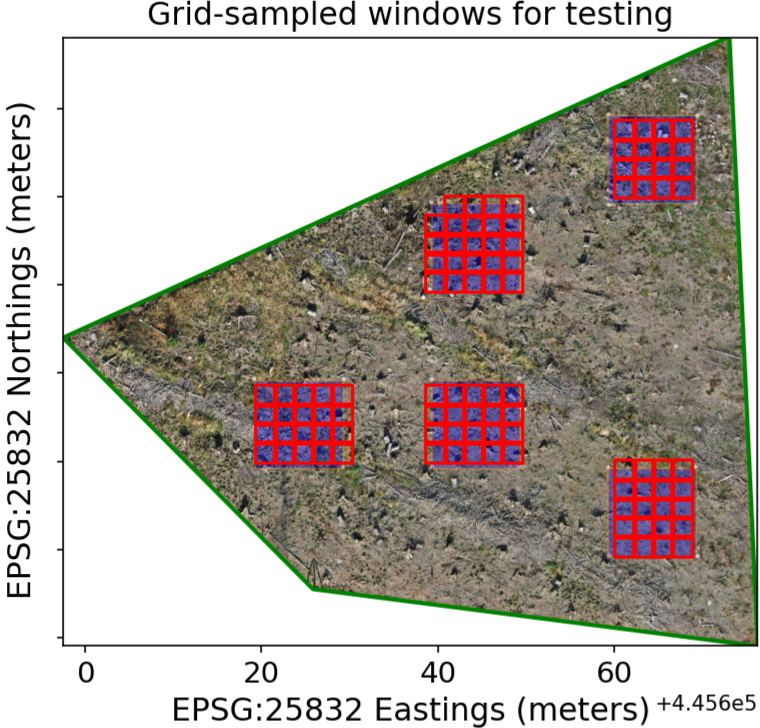} 
	\caption{Example of grid-sampled windows for the test set, sampled from the tiles designated for testing.}
	\label{fig:grid_sampled_windows_for_test_set}
\end{figure}

To facilitate the evaluation and benchmarking of remote sensing (RS) datasets, we developed the Garrulus dataset (GDL) library \footnote{\href{https://github.com/garrulus-project/gdl}{https://github.com/garrulus-project/gdl}}.
The dataset was divided into training and testing splits using 10$\times$10 m tiles, with windows sampled from their respective tiles. To ensure data validity, sampled windows were constrained to the dataset's spatial boundary, excluding no-data areas. All windows were extracted at a resolution of 512$\times$512 pixels. For training, 1000 samples were randomly generated from the training tiles, as illustrated in Fig.~\ref{fig:randomly_sampled_windows_for_train_set}. In contrast, the test set was created using a fixed grid sampling approach, resulting in 36 test samples, shown in Fig.~\ref{fig:grid_sampled_windows_for_test_set}. This number was chosen as it covers most of the areas within both train and test tiles. However, the number of samples can be increased depending on the size of the area. 
This approach provides flexibility that enables the incorporation of tiles from different fields into either training or testing sets. Furthermore, data augmentations such as random flip and rotation  are applied during training. Our code is available in our GitHub repository \footnote{\href{https://github.com/garrulus-project/sam_peft}{https://github.com/garrulus-project/sam\_peft}}.

\subsection{Experimental Setup}

The model was trained for 50 epochs with a batch size of 4 using stochastic gradient descent (SGD) as optimizer. The initial learning rate was set to 0.005, with a momentum of 0.9 and weight decay of 0.001. A warm-up strategy was used, in which the learning rate increased from 0 to the initial rate during the first 10\% of the total iterations, followed by a cosine annealing schedule that decreased the rate by a factor of 0.1 at each step. The loss function was a combination of weighted cross-entropy loss and Dice loss.

Two adapter modules, \textbf{AdapterH}  \cite{wu2023medicalsamadapter} and \textbf{AdapterL}\cite{chen2022adaptformer}, were evaluated, both with a bottleneck layer dimension of 32 and a scaling factor of 0.1. 
AdapterL has lower number of parameters as it only add the adapter module after the MLP layer. 
\textbf{LoRA} was implemented applying trainable matrices to the weights of the query $W_q$ and the value $W_v$ of each transformer layer in the image encoder and a rank of four, following the experimental setup conducted by Hu et al.~\cite{hu2022lora}.
Additionally, we fine-tuned the \textbf{SAM mask decoder} only, thus, not adding PEFT modules to the image encoder. All experiments were repeated three times with different random seeds, and the precision, recall, Dice score, and mean intersection over union (mIoU) were reported.

\subsection{Results and Discussion}

The quantitative performance of different PEFT methods is presented in Table \ref{tab:results}. We observe an mIoU of $0.416, 0.359, 0,410, 0.321$ for AdapterH, AdapterL, LoRA, and the SAM decoder respectively. The superiority of AdapterH over the other PEFT methods can also be seen from the precision, recall and Dice score. Although AdapterH achieved the highest score in all metrics, its improvement over LoRA is marginal considering that it doubles the number of its parameters. The performance of AdapterL, which injects parameters into the MLP layers only, is diminished across all metrics. Although AdapterL has a higher number of parameters compared to LoRA, it does not result in higher performance. This indicates that the location within the architecture, where PEFT modules are applied, does influence the results. Furthermore, fine-tuning only the SAM mask decoder, while leaving the SAM image encoder unchanged, yielded the lowest performance, suggesting the need to apply PEFT methods in the image encoder.


\begin{table}[htbp]
	\caption{Performance of PEFT Methods, pretraining on SAM mask decoder, and Zero-Shot SAM}
	\begin{center}
		\begin{tabular}{|c|c|c|}
			\hline
			\textbf{Method} & \textbf{Trainable Params} & \textbf{Metrics} \\
			\hline
			AdapterH & 9.66M & Precision: 0.827 $\pm$ 0.004 \\ 
			&       & Recall: \textbf{0.717} $\pm$ 0.002 \\  
			&       & Dice: \textbf{0.897} $\pm$ 0.000 \\  
			&       & mIoU: \textbf{0.416} $\pm$ 0.001 \\
			\hline
			AdapterL & 7.00M & Precision: 0.714 $\pm$ 0.078 \\  
			&       & Recall: 0.580 $\pm$ 0.020 \\  
			&       & Dice: 0.863 $\pm$ 0.001 \\  
			&       & mIoU: 0.359 $\pm$ 0.010 \\
			\hline
			LoRA & 4.99M & Precision: \textbf{0.830} $\pm$ 0.002 \\  
			&       & Recall: 0.694 $\pm$ 0.003 \\  
			&       & Dice: 0.894 $\pm$ 0.001 \\  
			&       & mIoU: 0.410 $\pm$ 0.001 \\
			\hline
			SAM Decoder & 4.34M & Precision: 0.598 $\pm$ 0.079 \\  
			&       & Recall: 0.545 $\pm$ 0.017 \\  
			&       & Dice: 0.824 $\pm$ 0.003 \\  
			&       & mIoU: 0.321 $\pm$ 0.010 \\
			\hline
			SAM (zero-shot) & 0M & mIoU: 0.151 \\
			\hline
		\end{tabular}
		\label{tab:results}
	\end{center}
\end{table}

Zero-shot experiments were conducted using SAM without incorporating PEFT or fine-tuning its mask decoder.
For mask generation, we utilized SAM’s automatic mask generator with point prompts, following the approach by Hu et al.~\cite{hu2022lora}. Specifically, each sampled window from the orthomosaic was propagated to SAM along with a 32×32 grid of point prompts, where each point produced a set of candidate masks representing potential objects. To refine the predictions, confidence masks were then selected based on their IoU scores. As these point prompts can lead to overlapping mask predictions, non-maximum suppression (NMS) is applied to eliminate redundancies.

Since SAM does not assign categories to masks, we employed a thresholding strategy, considering a predicted mask as a true positive if its IoU with the ground truth mask exceeded 50\%. The results on this zero-shot experiment, as presented in Table \ref{tab:results}, demonstrate a significantly low mean IoU. Particularly, SAM struggles to accurately segment vegetation where the object boundaries often lack clear separation from the background. This result aligns with previous studies \cite{wu2023medicalsamadapter,ge2024fine-tuning,chen2024rsprompter} that demonstrate fine-tuning for specific datasets.

\begin{figure}[!htbp]
	\centering
	\includegraphics[width=\linewidth]{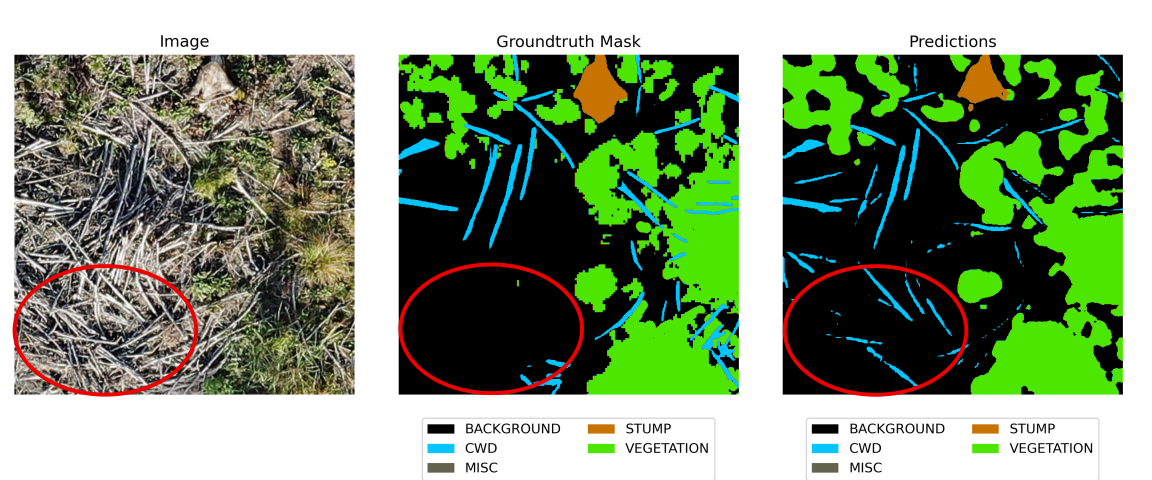} 
	\caption{Labeling inconsistencies in the CWD class lead to a high number of false positives. The red-circled area contains numerous unlabeled woody debris.}
	\label{fig:adapter_h_cwd_fp}
\end{figure}

\begin{figure}[!htbp]
	\centering
	\includegraphics[width=\linewidth]{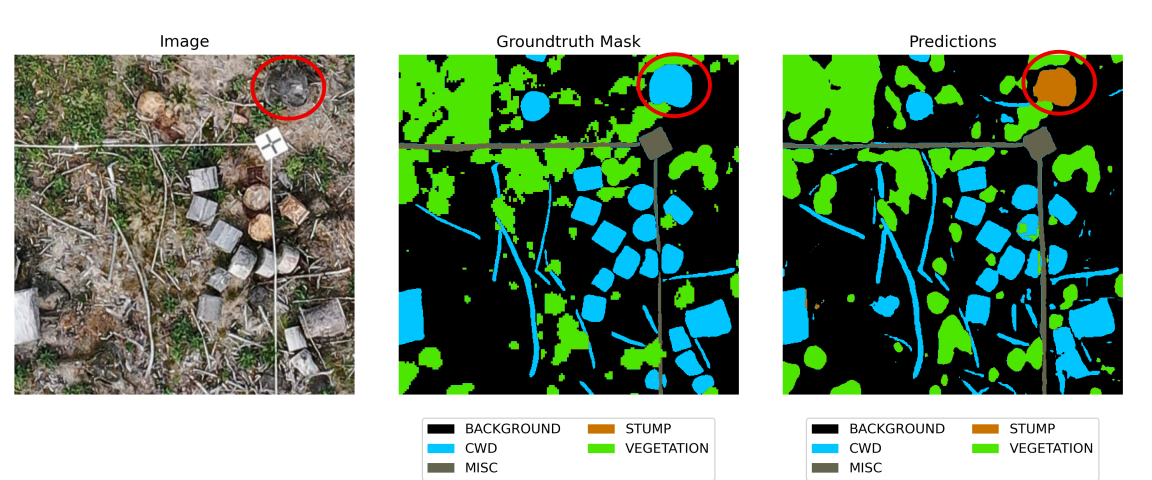} 
	\caption{CWD class labeling issues (highlighted in red), in which chopped upside-down logs are labeled as woody debris. However, the model predicts it as a tree stump due to their similar features.}
	\label{fig:cwd_labeling_issues}
\end{figure}

\begin{figure}[!htbp]
	\centering
	\includegraphics[width=\linewidth]{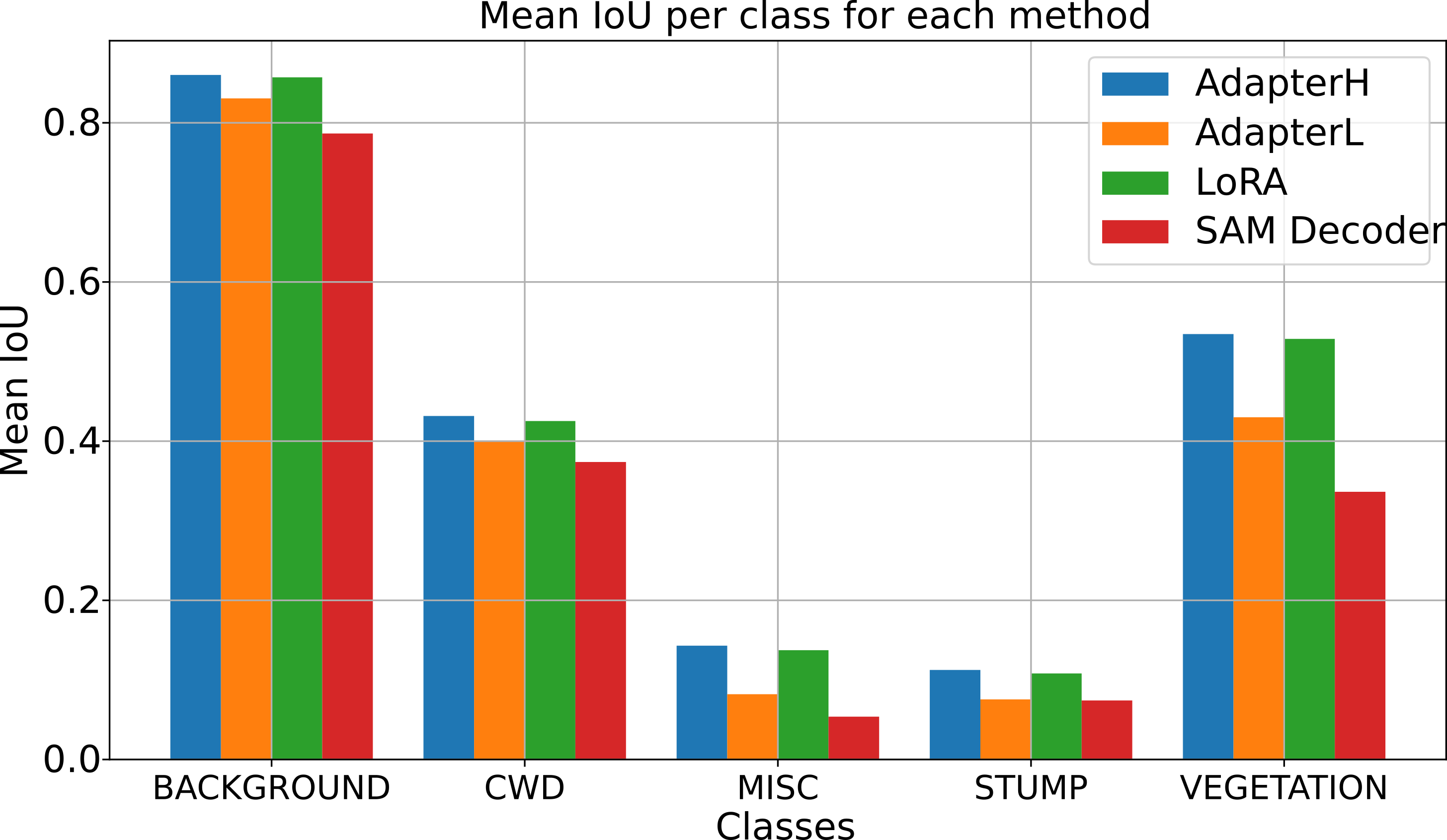} 
	\caption{Per-class mIoU, showing that performance is dominated by the BACKGROUND class, while the MISC and STUMP classes have significantly lower mIoU.}
	\label{fig:miou_per_class}
\end{figure}

We observed that, across all PEFT methods, the recall performance was consistently lower than the precision. This discrepancy can be partially attributed to labeling inconsistencies within the Garrulus dataset. For example, many instances of coarse woody debris (CWD) are incorrectly labeled or entirely missed. Furthermore, the subjective nature of what constitutes CWD, as illustrated in Fig. \ref{fig:cwd_labeling_issues}, adds to the problem. Specifically, an upside-down log, albeit labeled as CWD, shares visual similarities with the STUMP class. These labeling issues contribute to higher false positive rates in model predictions, as demonstrated in Fig. \ref{fig:adapter_h_cwd_fp}. This is also reflected in the per-class mIoU values, where CWD consistently shows the lowest mIoU, as depicted in Fig. \ref{fig:miou_per_class}.

\begin{figure}[!htbp]
	\centering
	\includegraphics[width=\linewidth]{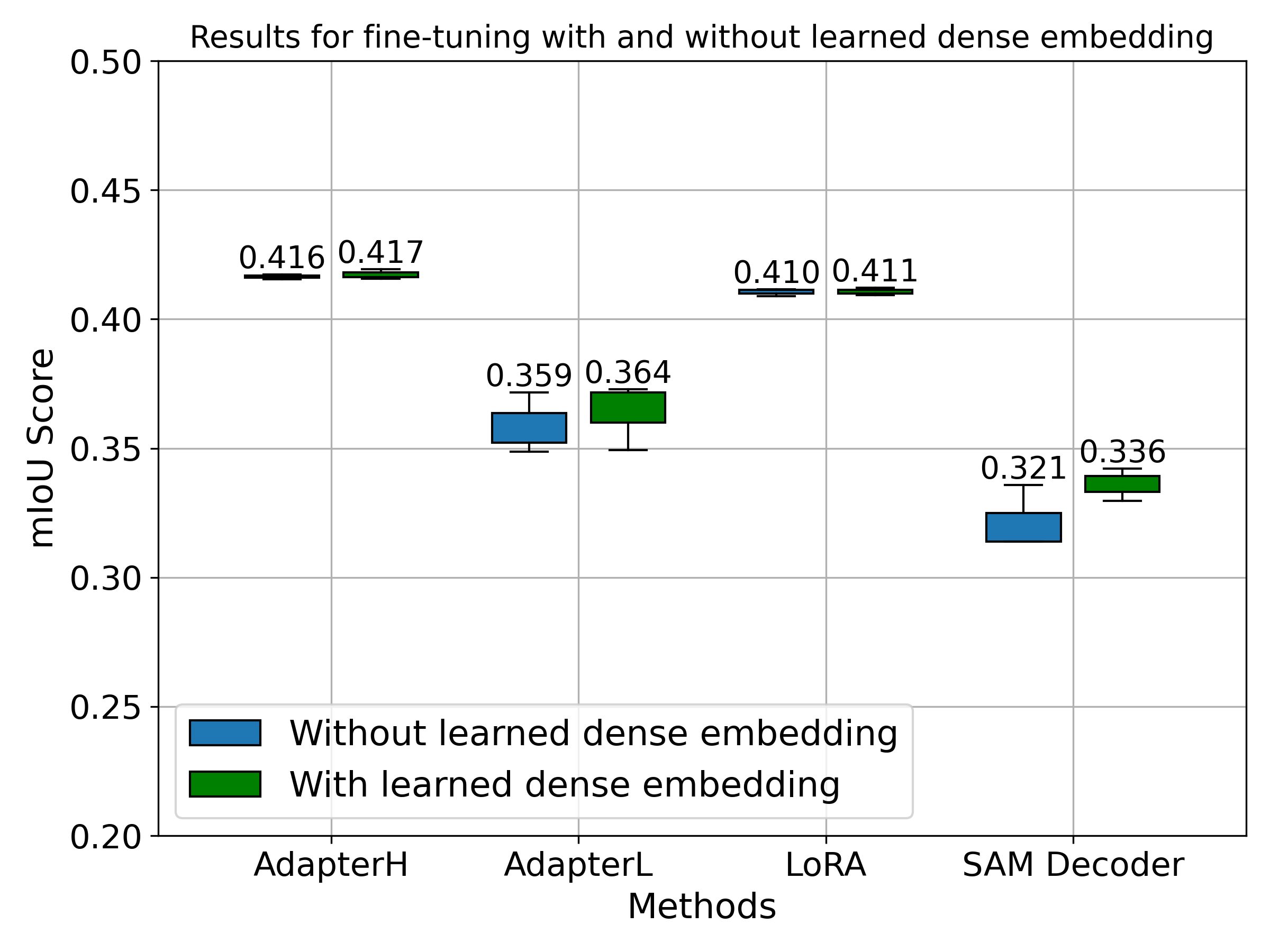} 
	\caption{mIoU performance comparison between fine-tuning with and without learned dense embedding. The value above the box plot shows the mean calculated from different seeds.}
	\label{fig:miou_dense_wo_dense_embed}
\end{figure}

We conducted an ablation study to analyze the impact of adding a learned dense embedding (\textit{no-mask embedding}) to the image embedding. This dense embedding is originally used by SAM if there is no mask prompt provided, and introduces an additional 6220 trainable parameters. Without the learned dense embedding, the mask decoder input solely relies on the image embedding. We observed that adding this learned dense embedding does not significantly improve the performance when usign AdapterH and LoRA while it introduces higher computation and memory requirements as shown in Fig. \ref{fig:miou_dense_wo_dense_embed}. However, for AdapterL and the SAM decoder, it shows an mIoU improvement of 0.02\% and 0.04\% respectively.

\section{Conclusion and Future Work}

We have demonstrated the adaptation of the well-known Segment Anything Model (SAM) using Parameter-Efficient Fine-Tuning (PEFT) methods for segmenting forest floor objects.
By removing SAM's prompt encoder and modifying its mask decoder to output mask tokens corresponding to the number of classes in the dataset, the model can perform automatic segmentation without requiring additional prompts.
The results indicate that adapter- and LoRA-based fine-tuning outperform direct fine-tuning of SAM's mask decoder. Additionally, zero-shot inference using SAM on the forest floor dataset resulted in a significantly low mIoU. Among the PEFT methods, AdapterH (serial adapter) achieved the highest mIoU, while LoRA, despite slightly lower performance, required only half the parameters of AdapterH. This makes LoRA a more efficient choice for deployment on resource-constrained UAV platforms. 

For future work, we will adapt the trained PEFT models to diverse environmental conditions, such as on-the-fly adaptation to different locations, weather or environmental conditions. Additionally, exploring real-time distillation of a foundation model and investigating the feasibility of online training directly on UAV platforms are promising directions for further research. Furthermore, we aim to integrate the identification of obstacle-free areas, based on SAM, into our seeding pipeline.

\section*{Acknowledgment}

This work is conducted in the context of the Garrulus project, which is funded by the Ministry for Agriculture and Consumer Protection of the State of North Rhine-Westphalia Germany. This work has been supported by the Bonn-Aachen International Center for Information Technology,  and the Graduate Institute at Hochschule Bonn-Rhein-Sieg.


\bibliographystyle{IEEEtran}
\bibliography{bibliography.bib}

\end{document}